\documentclass{article}

\usepackage{PRIMEarxiv}
\usepackage[utf8]{inputenc}
\usepackage[T1]{fontenc}
\usepackage{hyperref}
\usepackage{url}
\usepackage{booktabs}
\usepackage{amsfonts}
\usepackage{nicefrac}
\usepackage{microtype}
\usepackage{lipsum}
\usepackage{fancyhdr}
\usepackage{graphicx}
\usepackage{listings}
\graphicspath{{media/}}
\usepackage{caption}
\usepackage{subcaption}
\usepackage{amsmath}

\title{Decision Tree Psychological Risk Assessment in Currency Trading

}

\author{
  Jai Pal \\
  Independent Researcher \\
  \texttt{jaipal9621@gmail.com}
}

\begin{document}
\maketitle

\begin{abstract}
This research paper focuses on the integration of Artificial Intelligence (AI) into the currency trading landscape, positing the development of personalized AI models, essentially functioning as intelligent personal assistants tailored to the idiosyncrasies of individual traders. The paper posits that AI models are capable of identifying nuanced patterns within the trader's historical data, facilitating a more accurate and insightful assessment of psychological risk dynamics in currency trading. The PRI is a dynamic metric that experiences fluctuations in response to market conditions that foster psychological fragility among traders. By employing sophisticated techniques, a classifying decision tree is crafted, enabling clearer decision-making boundaries within the tree structure. By incorporating the user's chronological trade entries, the model becomes adept at identifying critical junctures when psychological risks are heightened. The real-time nature of the calculations enhances the model's utility as a proactive tool, offering timely alerts to traders about impending moments of psychological risks. The implications of this research extend beyond the confines of currency trading, reaching into the realms of other industries where the judicious application of personalized modeling emerges as an efficient and strategic approach. This paper positions itself at the intersection of cutting-edge technology and the intricate nuances of human psychology, offering a transformative paradigm for decision making support in dynamic and high-pressure environments.
\end{abstract}

\section{Introduction}
In the fast-paced and highly competitive realm of currency trading, individuals engaging in this financial pursuit often find themselves navigating through a landscape fraught with psychological pressures, such as the allure of greed, the paralyzing grip of fear, and the unsettling presence of doubt. For both seasoned professionals and novice traders, these emotional challenges can significantly impact decision-making, potentially leading to detrimental trading patterns.

Recognizing the inherent difficulty of managing these psychological dynamics, there is a growing interest in leveraging Artificial Intelligence (AI) to assist traders in enhancing their decision-making processes. The concept involves the creation of personalized AI models designed to function as vigilant personal assistants. These AI companions play a crucial role in identifying and alerting traders to moments of psychological and emotional fragility, which have been known to trigger risky and destructive trading behaviors.

To achieve this, the AI models utilize a trader's personal journal—a rich source of information documenting past trades, emotional states, and decision-making processes. By employing sophisticated techniques, a classifying decision tree is crafted, capable of identifying nuanced patterns within the trader's historical data. This personalized approach enables the AI to deliver tailor-made insights and preemptively intervene to prevent the recurrence of dangerous trading practices.

The integration of AI in the currency trading landscape not only aims to enhance decision-making but also to foster a more disciplined and resilient approach to trading. By offering personalized insights derived from an individual trader's experiences, the AI becomes an invaluable ally, empowering traders to make informed decisions, manage emotions effectively, and navigate the complexities of currency markets with greater confidence and success.

\section{Data Collection and Cleaning}
\subsection{Personal Journals}
In developing a personalized AI model for trading, the individual trader's personal journal is a crucial data source. This journal, containing details like account balances, risk-to-reward ratios, and trade outcomes, provides a unique and intimate account of the trader's strategies and emotional responses to market changes. Despite its seemingly basic nature, this data offers valuable insights into a trader's decision-making and risk management.

Analyzing past trades can reveal patterns indicative of a trader's response to stress and uncertainty. Additionally, risk-to-reward ratios highlight their approach to risk management. This personalized data, supported by existing research in the psychology of money, allows the AI model to make informed predictions about a trader's psychological strengths and weaknesses.

By leveraging this wealth of information, the AI model tailors its interventions, offering nuanced and personalized support aligned with each trader's unique profile.

\begin{lstlisting}[language=Python, caption={Selecting Columns from Journal}, label={lst:python}]
df = pd.read_csv('log.csv')
df = df[['Max RR', 'Rs', 'BE', 'Session']]
df['BE'] = df['BE'].apply(lambda x: 1 if x == "W" else -1)
\end{lstlisting}

\subsection{PRI Calculation and Variability}
The Psychological Risk Index (PRI) is a dynamic metric that experiences fluctuations in response to market conditions that foster psychological fragility among traders. It's important to recognize the substantial variation in PRI across individual traders, and this diversity can be attributed to the inherent differences in people's subconscious attitudes toward money and their unique perceptions of their environment.

The rationale behind this variability is straightforward: individuals bring distinct psychological predispositions to their trading activities, influenced by personal experiences, risk tolerance, and cognitive biases. These differences manifest in the way traders assess and respond to market dynamics. Consequently, the PRI's formulation is not a one-size-fits-all model; instead, it adapts to the idiosyncrasies of each trader's psychological makeup.

In the context of this research paper, a specific formula for the PRI is devised to accurately mirror the characteristics observed in the sample trading journal. While this formula may be considered arbitrary for the broader trading community, it is meticulously tailored to align with the intricacies of the studied dataset, ensuring a faithful representation of the psychological risk dynamics inherent in the observed trading behaviors. In a real world application, a formula can also be disregarded and the trader can manually review their psychological fragility in historical trades.

\begin{lstlisting}[language=Python, caption={Calculating PRI}, label={lst:python}]
# Update 'PRI' based on specified conditions
    if abs(df.at[index, 'Streak']) >= 3:
        df.at[index, 'PRI'] += 1
    
    if index >= 3 and (df.at[index - 1, 'Max RR'] / 3 > 7.5):
        df.at[index, 'PRI'] += 1
    
    # Calculate historical session losses for all sessions
    session_columns = [col for col in df.columns if col.startswith('Session_')]
    historical_losses = {session: df[df[session] == 1]['BE'].eq(-1).sum()
     for session in session_columns}
    
    # Find the session with the most losses historically
    max_loss_session = max(historical_losses, key=historical_losses.get)
    
    # Check if the current trade is in the session with the most losses historically
    if row[max_loss_session] == 1:
        df.at[index, 'PRI'] += 1
\end{lstlisting}

\subsection{Pruning Data}
In the process of extracting meaningful insights from trading journals, it becomes evident that a substantial portion of the information, such as subjective comments and reviews, poses challenges for accurate interpretation by AI models. To address this inherent subjectivity, a careful pruning of the journals is undertaken, focusing on extracting specific and quantifiable data points that are conducive to objective analysis.

From the sample journal, key data columns are selected to construct a comprehensive yet focused dataset. These columns include the maximum risk-to-reward ratio (RR), success, streak, session, and balance. The maximum risk-to-reward ratio serves as a crucial metric, providing insights into the trader's strategic aggression. Higher risk-to-reward ratios often correlate with unstable psychological complexes, reflecting a heightened level of risk tolerance and potential emotional volatility.

The success column delineates the outcome of a trade entry, indicating whether profits are secured or not. Streak, on the other hand, captures the consecutive wins and losses, resetting when this pattern is disrupted. Extended winning streaks tend to induce greed, thereby amplifying psychological risk. The session column, representing the time block during which a trade occurs, is instrumental in assessing the impact of trading activities within specific temporal contexts. A success rate below 50\% in a single session is presumed to depreciate morale, contributing to an increased sense of desperation, which correlates with higher psychological risk.

Additionally, the balance column is calculated hypothetically for the sample. This balance metric assumes a hypothetical scenario, evaluating the trader's financial standing. Similar to the success rate, a low or negative net balance is indicative of financial desperation, aligning with higher psychological risk. This meticulous selection of columns ensures that the AI model is fed with data points that encapsulate critical aspects of a trader's behavior, allowing for a more nuanced and accurate assessment of psychological risk dynamics.

\begin{lstlisting}[language=Python, caption={Final Data Categories}, label={lst:python}]
df = pd.read_csv('clean_log.csv')
features = ['Max RR', 'BE', 'Streak', 'Session_Asian', 'Session_London']
target = 'PRI'
\end{lstlisting}

\subsection{Use of Hypothetical Balance}
Upon the completion of the model, it became apparent that the inclusion of the hypothetical balance as a parameter had unintended consequences. Instead of enhancing the model's predictive capabilities, the hypothetical balance emerged as a distracting filler parameter, occupying the pivotal position of the root node. This inadvertent inclusion led to an unnecessary increase in the tree depth across all trials in which it was incorporated.

Recognizing the need for a more streamlined and focused decision tree, the decision was made to remove the hypothetical balance from the set of parameters. This strategic adjustment aimed to guide the model towards prioritizing more significant and holistic indicators that align with the predetermined formula for the Psychological Risk Index (PRI). By eliminating the distracting influence of the hypothetical balance, the model could more effectively discern and emphasize the key features and patterns within the data, ultimately contributing to a more accurate and insightful assessment of psychological risk dynamics in currency trading.

\subsection{Binarizing PRI}
In applying the PRI algorithm, the resulting values can range from zero to three, representing varying degrees of psychological risk. However, upon examining the sample data, it was noted that PRI values exceeding 2 were not observed. Despite this, the decision was made to categorize the PRI into three distinct outcomes: zero, one, and two. This binarization process aimed to align the PRI values with the inherently binary structure of the decision tree.

Binarization involves transforming the PRI values into a format that fits seamlessly into the decision tree's binary framework. By categorizing the PRI into discrete outcomes, the model could effectively integrate this crucial psychological risk indicator, facilitating clearer decision-making boundaries within the tree structure. This strategic approach not only streamlined the model's interpretation but also ensured compatibility with the binary nature of decision tree algorithms, enhancing the overall effectiveness of the PRI in predicting and managing psychological risk in currency trading.

\begin{lstlisting}[language=Python, caption={Binarizing PRI}, label={lst:python}]
y_bin = label_binarize(y_test, classes=[0, 1, 2])
classifier = OneVsRestClassifier(best_model)
classifier.fit(X_train, label_binarize(y_train, classes=[0, 1, 2]))
\end{lstlisting}

\section{Model Selection}
When considering the primary contenders in machine learning models, namely neural networks and decision trees, the pivotal factor influencing the choice lies in the nature of the available data. Given the requirement for personalization in the model, data scarcity becomes a significant challenge. In this context, neural networks face limitations as they demand substantial amounts of data to perform optimally, making them less suitable for this specific project. In order to split the data at the tree nodes, the Gini index is used.

The Gini index is defined as
\[
\text{Gini}(p) = 1 - \sum_{i=1}^{K} p_i^2,
\]
where \(p\) is the vector of class probabilities, and \(K\) is the number of classes.

Moreover, the architectural complexity of neural networks poses another hurdle, particularly in terms of human interpretability. Decision trees, on the other hand, offer a more transparent and comprehensible structure. This characteristic becomes crucial in scenarios where users may wish to review the PRI calculation. The simplicity of visualizing a decision tree allows users to grasp the model's decision-making process with ease.

Considering the constraints of data scarcity and the need for interpretability, the decision tree emerges as the optimal choice for this project. Its capacity to handle personalized data effectively and provide human-understandable insights positions it as the most suitable model for constructing a robust Psychological Risk Index in the context of currency trading.

\begin{figure}[ht]
    \centering
    \includegraphics[width=1\linewidth]{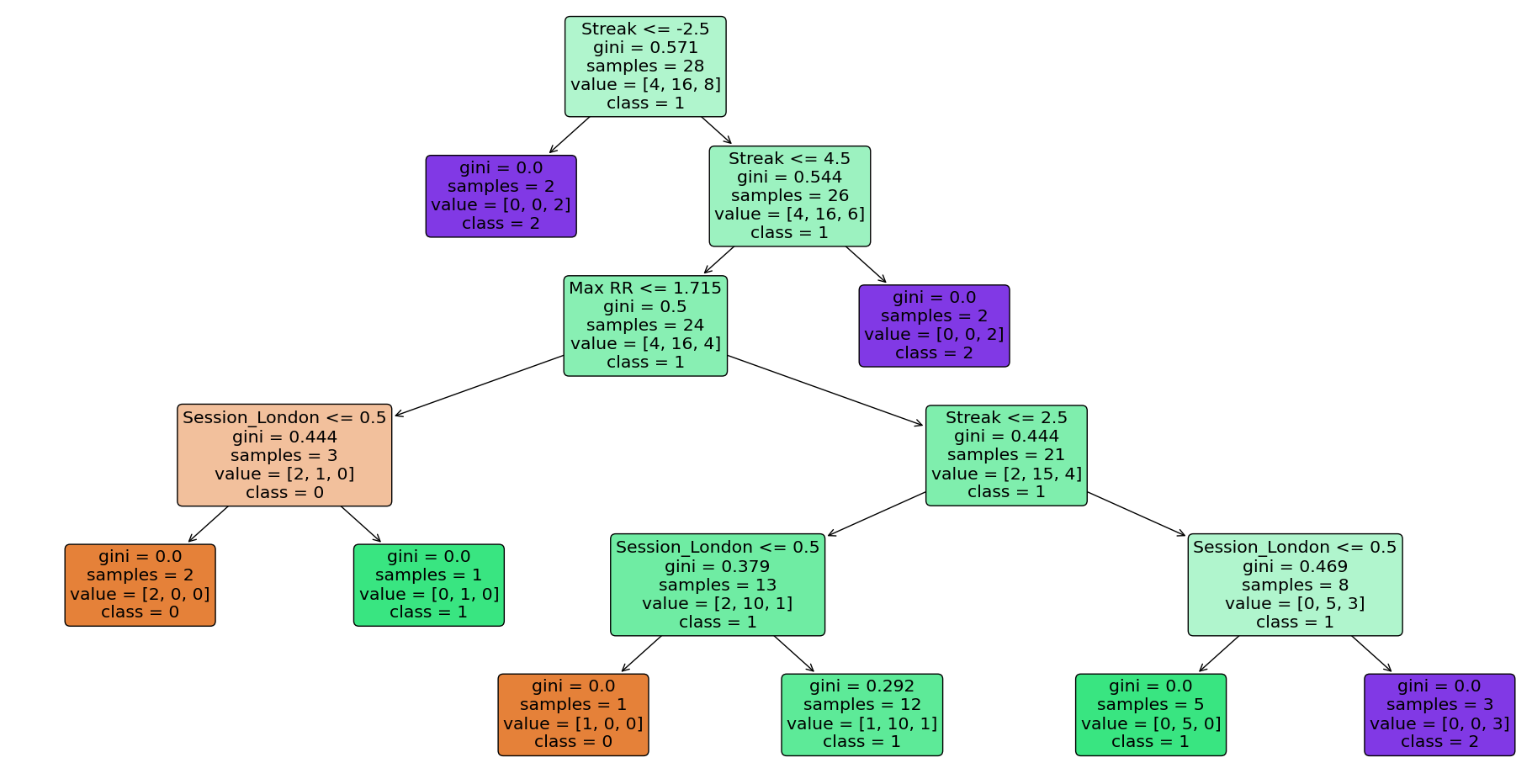}
    \caption{Decision Tree Visualization}
    \label{fig:enter-label}
\end{figure}

\section{Parameter Selection}
In the pursuit of identifying the optimal parameters for achieving both accuracy and conciseness in the decision tree model, a systematic approach was employed using a parameter grid. The three key parameters under scrutiny were the maximum tree depth, minimum samples per leaf, and minimum samples per split. This grid was subjected to rigorous testing, allowing for an exhaustive exploration of various parameter combinations.

The objective was to automatically pinpoint the parameter values that resulted in the most accurate predictions while maintaining a concise tree structure. In the case of the sample trading journal, the experiments consistently revealed that a maximum tree depth of 5, a minimum samples per leaf of 1, and a minimum samples per split of 2 yielded the most accurate and efficient outcomes.

This meticulous parameter tuning process ensures that the decision tree is finely tuned to the specific characteristics of the individual's trading journal. By systematically assessing the impact of different parameter configurations, the model adapts to the nuances of the data, striking a balance between accuracy and simplicity in its predictive capabilities.

\begin{lstlisting}[language=Python, caption={Creating Test Parameter Grid}, label={lst:python}]
param_grid = {
    'max_depth': [3, 5, 7],
    'min_samples_split': [2, 5, 10],
    'min_samples_leaf': [1, 2, 4],
}
\end{lstlisting}

\begin{figure}
    \centering
    \includegraphics[width=1\linewidth]{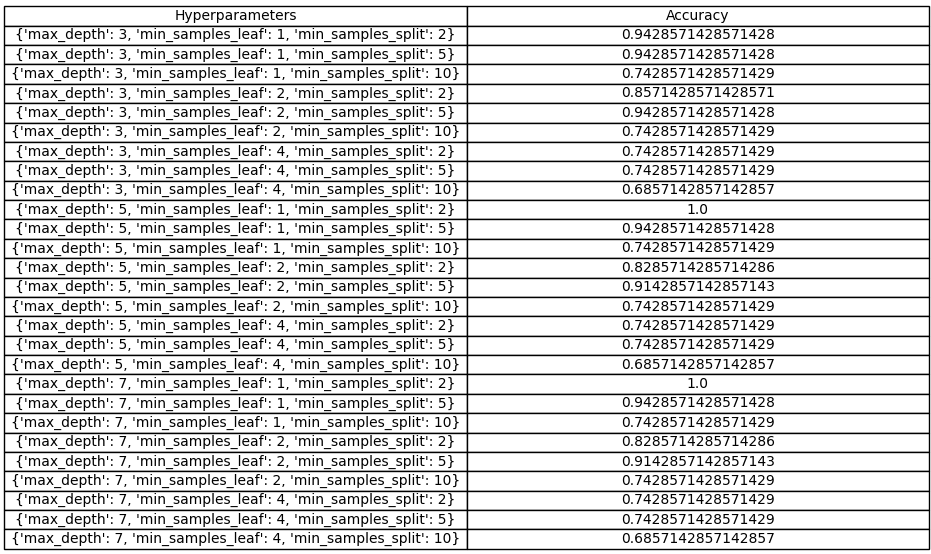}
    \caption{Parameter Accuracies}
    \label{fig:enter-label}
\end{figure}

\begin{lstlisting}[language=Python, caption={Testing Parameters}, label={lst:python}]
for params in ParameterGrid(param_grid):
    # Initialize an empty list to store predictions for the current hyperparameters
    predictions = []

    # Iterate through the dataset in chronological order
    for i in range(len(df)):
        # Use only past trades for training
        train_data = df.loc[:i, features]
        train_target = df.loc[:i, target]

        # Train a Decision Tree model with current hyperparameters
        model = DecisionTreeClassifier(random_state=42, **params)
        model.fit(train_data, train_target)

        # Make a prediction for the current trade
        current_trade_data = df.loc[i, features].values.reshape(1, -1)
        prediction = model.predict(current_trade_data)

        # Store the prediction
        predictions.append(prediction[0])

    # Add the predictions to the list of all predictions
    all_predictions.append({'params': params, 'predictions': predictions})

    # Evaluate the model with the current hyperparameters
    accuracy = accuracy_score(df[target], predictions)
    print(f"Accuracy for hyperparameters {params}: {accuracy}")

    # Store hyperparameters and accuracy
    hyperparameters.append(params)
    accuracies.append(accuracy)
\end{lstlisting}

\section{Iterative Training Style}
The uniqueness of the model's training style is a pivotal aspect of its effectiveness. Given the chronological organization of the trading journals and the necessity to maintain this temporal structure, the model adopts an iterative training approach. This distinctive method involves feeding trade entries to the model one at a time, with each prediction taking into account the evolving context of the preceding trades.

This iterative training approach holds a distinct advantage over holistic training paradigms, where all data is considered collectively. By assimilating information incrementally, the model is compelled to adapt in real-time to the evolving dynamics of the trader's decision-making process. This dynamic evolution not only mirrors the psychological progression of the trader but also opens avenues for real-time applications, allowing the model to respond promptly to changes in the trader's behavior and market conditions. The iterative training style enhances the model's responsiveness and relevance, making it a valuable tool for dynamic trading scenarios.

\begin{lstlisting}[language=Python, caption={Iterative Training Approach}, label={lst:python}]
for params in ParameterGrid(param_grid):
    # Initialize an empty list to store predictions for the current hyperparameters
    predictions = []

    # Iterate through the dataset in chronological order
    for i in range(len(df)):
        # Use only past trades for training
        train_data = df.loc[:i, features]
        train_target = df.loc[:i, target]

        # Train a Decision Tree model with current hyperparameters
        model = DecisionTreeClassifier(random_state=42, **params)
        model.fit(train_data, train_target)

        # Make a prediction for the current trade
        current_trade_data = df.loc[i, features].values.reshape(1, -1)
        prediction = model.predict(current_trade_data)

        # Store the prediction
        predictions.append(prediction[0])

    # Add the predictions to the list of all predictions
    all_predictions.append({'params': params, 'predictions': predictions})
\end{lstlisting}

\section{Performance}
The model exhibited outstanding proficiency in the calculation of the Psychological Risk Index, showcasing remarkable accuracy. Leveraging the parameters that yielded the highest accuracy, the F1 score, Precision, and Recall all achieved a flawless score of 1.0. This exceptional alignment between predicted and actual values underscores the model's precision and effectiveness in identifying instances of psychological risk with unparalleled accuracy. The convergence of these evaluation metrics at their maximum values signifies not only the model's ability to avoid false positives and false negatives but also its capacity to comprehensively capture and classify instances of psychological fragility.

Furthermore, to visually represent the model's performance, a Receiver Operating Characteristic (ROC) Curve has been provided. The ROC Curve serves as a valuable tool in assessing the model's discriminative ability across different thresholds. In the context of this study, the ROC Curve enhances the interpretability of the model's predictive performance, offering insights into its ability to distinguish between positive and negative instances. The inclusion of this curve further contributes to the comprehensive evaluation of the model's robustness and its capacity to provide accurate psychological risk predictions in the realm of currency trading.

\begin{figure}[h]
    \centering
    \begin{subfigure}[b]{0.3\textwidth}
        \includegraphics[width=\textwidth]{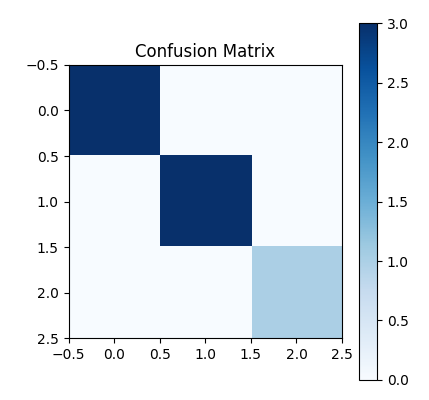}
        \caption{Confusion Matrix}
        \label{fig:figure1}
    \end{subfigure}
    \hspace{1cm} % Adjust the horizontal space between figures
    \begin{subfigure}[b]{0.3\textwidth}
        \includegraphics[width=\textwidth]{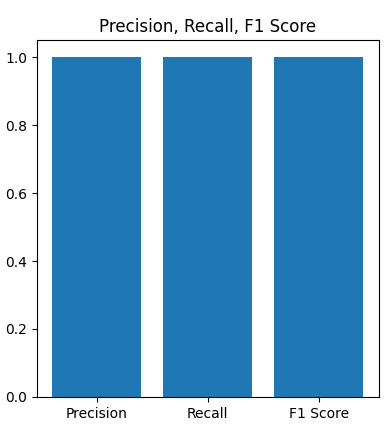}
        \caption{Accuracy Metrics}
        \label{fig:figure2}
    \end{subfigure}
    \hspace{1cm} % Adjust the horizontal space between figures
    \begin{subfigure}[b]{0.3\textwidth}
        \includegraphics[width=\textwidth]{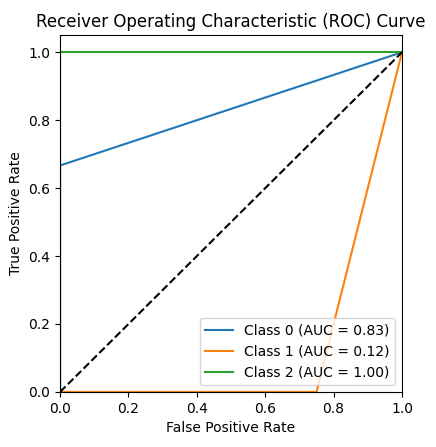}
        \caption{ROC Curve}
        \label{fig:figure3}
    \end{subfigure}
    \caption{Performance Metrics}
    \label{fig:overall}
\end{figure}

\section{Shortcomings}
One notable limitation of the model lies in its individualized training requirement for each trader. Unlike models that can leverage publicly available data, this approach demands that users contribute their personal trading journals, creating a unique dataset for each individual. While this personalized approach enhances the model's accuracy for the specific user, it introduces challenges related to biases and potential inconsistencies in the provided data.

However, it is crucial to recognize that, in this context, the model's sensitivity to biases can be viewed as a positive attribute. The emphasis on personalization necessitates the model to adapt to the specific nuances of an individual trader's behavior. In this light, the model's ability to fine-tune itself based on the user's unique trading patterns and psychological attributes aligns with the overarching goal of tailoring the AI to the user's specific needs and tendencies. Although the need for individualized training poses a logistical challenge, it also contributes to the model's effectiveness in providing personalized insights and risk assessments.

\section{Application}
The envisioned application for this research extends into the integration of psychological risk predictions into real-time trading tools. Leveraging the iterative training style, the model excels in contextualizing the entire trade history. This unique approach enables the model to perform real-time calculations, providing timely alerts to traders about impending moments of psychological fragility.

By incorporating the user's chronological trade entries and continuously evolving alongside the trader's decision-making patterns, the model becomes adept at identifying critical junctures when psychological risks are heightened. The real-time nature of the calculations enhances the model's utility as a proactive tool, offering traders timely insights into their psychological states during ongoing trading activities. This application holds the potential to significantly contribute to traders' decision-making processes, allowing for more informed and psychologically-aware trading strategies.

\section{Conclusion}
This research paper immerses itself in the complex landscape of currency trading, shedding light on the formidable psychological challenges that traders contend with, challenges that wield profound influence over their decision-making processes. At the core of this exploration is the innovative integration of Artificial Intelligence (AI) into the trading domain. The paper posits the development of personalized AI models, essentially functioning as intelligent personal assistants tailored to the idiosyncrasies of individual traders. These models stand as vigilant sentinels, adept at discerning and proactively addressing instances of psychological vulnerability among traders. The implications of this research extend beyond the confines of currency trading, reaching into the realms of real-time applications and other industries where the judicious application of personalized modeling emerges as an efficient and strategic approach. The paper positions itself at the intersection of cutting-edge technology and the intricate nuances of human psychology, offering a transformative paradigm for decision-making support in dynamic and high-pressure environments.

\section{Bibliography}
1.10. decision trees. (n.d.). Scikit-Learn. Retrieved November 26, 2023, from https://scikit-learn.org/stable/modules/tree.html\#classification

User Guide — pandas 2.1.3 documentation. (n.d.). Retrieved November 26, 2023, from https://pandas.pydata.org/docs/user\_guide/index.html\#user-guide

Using Matplotlib — Matplotlib 3.8.2 documentation. (n.d.). Retrieved November 26, 2023, from https://matplotlib.org/stable/users/index

\end{document}